\documentclass[%
reprint,
superscriptaddress,
amsmath,amssymb,
aps,
]{revtex4-1}

\usepackage{graphicx}
\usepackage{dcolumn}
\usepackage{bm}

\usepackage{todonotes} 

\begin{document}

\preprint{APS/123-QED}

\title{Boolean learning under noise-perturbations in hardware neural networks}

\author{Louis Andreoli}
\affiliation{FEMTO-ST/Optics Dept., UMR CNRS 6174, Univ. Bourgogne Franche-Comt\'{e}, 15B avenue des Montboucons, 25030 Besan\c{c}on Cedex, France}

\author{Xavier Porte}
\altaffiliation[]{Corresponding author: javier.porte@femto-st.fr}
\affiliation{FEMTO-ST/Optics Dept., UMR CNRS 6174, Univ. Bourgogne Franche-Comt\'{e}, 15B avenue des Montboucons, 25030 Besan\c{c}on Cedex, France}

\author{St\'ephane Chr\'etien}
\affiliation{Institut FEMTO-ST,  Universit\'e Bourgogne Franche-Comt\'e CNRS UMR 6174, Besan\c{c}on, France}

\author{Maxime Jacquot}
\affiliation{Institut FEMTO-ST,  Universit\'e Bourgogne Franche-Comt\'e CNRS UMR 6174, Besan\c{c}on, France}

\author{Laurent Larger}
\affiliation{Institut FEMTO-ST,  Universit\'e Bourgogne Franche-Comt\'e CNRS UMR 6174, Besan\c{c}on, France}

\author{Daniel Brunner}
\affiliation{Institut FEMTO-ST,  Universit\'e Bourgogne Franche-Comt\'e CNRS UMR 6174, Besan\c{c}on, France}

\date{\today}

\begin{abstract}
A high efficiency hardware integration of neural networks benefits from realizing nonlinearity, network connectivity and learning fully in a physical substrate. 
Multiple systems have recently implemented some or all of these operations, yet the focus was placed on addressing technological challenges. 
Fundamental questions regarding learning in hardware neural networks remain largely unexplored. 
Noise in particular is unavoidable in such architectures, and here we investigate its interaction with a learning algorithm using an opto-electronic recurrent neural network. 
We find that noise strongly modifies the system's path during convergence, and surprisingly fully decorrelates the final readout weight matrices. 
This highlights the importance of understanding architecture, noise and learning algorithm as interacting players, and therefore identifies the need for mathematical tools for noisy, analogue system optimization.
\end{abstract}

\maketitle

\section{Introduction} 

In recent years, neural networks (NNs) take centre-stage in advancing computation \cite{LeCun2015}.
Optimized by training, such \emph{learning} machines provide key advantages for solving abstract computational problems and already outperform humans in numerous tasks previously deemed impossible for classically (algorithmically) programmed computers \cite{LeCun2015,Amos2016,Graves2013}.

However, NNs are still mostly emulated by traditional Turing / von Neumann computers.
The absence of computing hardware supporting fully parallel neural networks reduces energy efficiency and overall speed, and new hardware paradigms addressing these problems are desirable.
An implementation of nonlinear neurons, fully-parallel information transduction and learning on a substrate level promises a revolution of today's neural network hardware, and photonic NNs \cite{Farhat1985a,Psaltis1985} remain a highly promising avenue \cite{Lin2018,Shen2016}. The lack of a parallel network substrate is a fundamental roadblock and is an active area of research since decades, with current analogue hardware either implementing the full network \cite{Lin2018,Shen2016,Tait2014} or the neurons \cite{Appeltant2011,Duport2012,Larger2012,Brunner2015,Torrejon2017NeuromorphicOscillators}.

Noise is an inseparable companion of analogue hardware \cite{Semenova2019}, yet the fundamental aspects of optimizing a noisy neural network \cite{Hermans2016,Antonik2017,Bueno2018} have so far hardly been explored - neither in experiments \cite{Alata2020,Soriano2013} nor in theory \cite{Semenova2019}.
Here, we investigate the interactions between noise, learning rules and the topology of an error landscape for the first time.
We experimentally implement a NN with 961 electro-optical neurons via a spatial-light modulator (SLM) \cite{Bueno2018}, use diffraction \cite{Maktoobi2020,Brunner2015,Lin2018,Shen2016} to physically realize the network's internal connections and a digital micro-mirror device (DMD) for programmable Boolean readout weights \cite{Bueno2018}.
Learning exclusively optimizes the readout connections \cite{Jaeger2004} via an evolutionary Boolean algorithm based on the error gradient only, and the error landscape's dimensions are probed according to either fully random (Markovian) or structured (greedy) exploration.

Learning trajectory statistics proof that noise and exploration strategy strongly interact.
Noise induces a kind of random forcing upon the descent algorithm, which strongly modifies the system’s path towards a local minimum.
We find that noise decorrelates the final weight configurations: starting from identical weight configurations and exploring the error landscape's dimensions in identical sequences always leads to clearly differentiated local minima.
Quite astonishingly, all minima are spaced at an almost constant distance from each other, which for the generally non-trivial error landscape topologies is unusual at the least.
Noise therefore appears to arrange minimizers in periodic positions, much like competitive Brownian walkers with non-local interactions \cite{Heinsalu2012}.
These fundamental interactions highlight the importance of considering hardware architecture, noise and learning algorithm as intimately linked.

\section{Neural network hardware}
\label{sec:NeuralNetwork}

\begin{figure}[t]
    \centering
    \includegraphics[width=0.95\columnwidth]{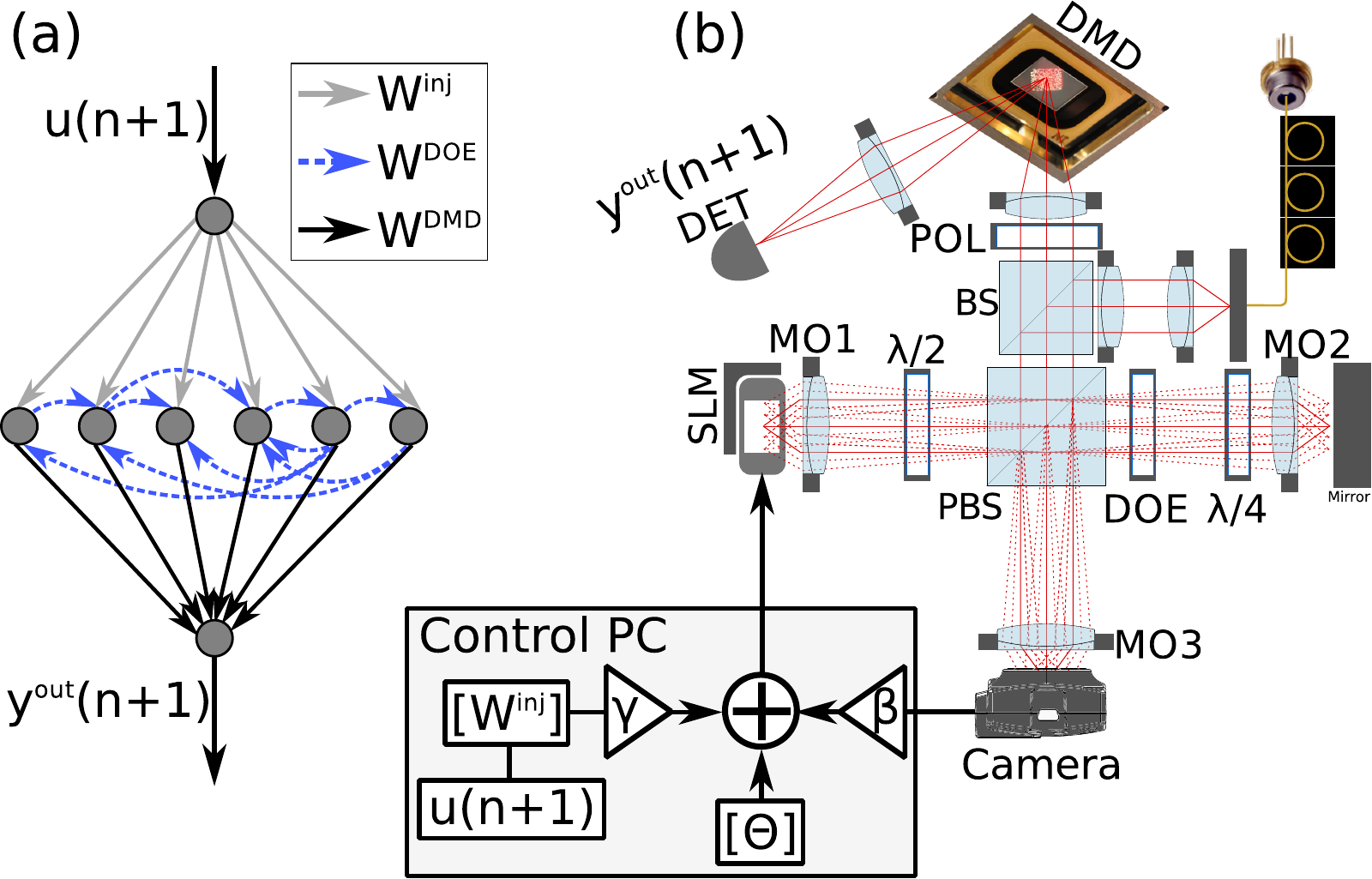}
    \caption{(a) Schematic illustration of a recurrent neural network.
        (b) Photonic implementation of a spatio-temporal neural network with 961 nodes.
        An optical plane wave illuminates the spatial light modulator (SLM), the neural network states are encoded by the SLM pixels.
        These are imaged on the camera, passing through a polarizing beam splitter (PBS) and the diffractive optical element (DOE) creating the coupling between network states.
        The information detected by the camera is used to drive the SLM.
        The network's output weights are realized via a digital micro-mirrors device (DMD) which creates Boolean readout weight matrix $\mathbf{W}^{\textrm{DMD}}$. }\label{fig:ExperimentalSetup}
    \hrule
\end{figure}

A recurrent neural network inspired by reservoir computing (RC), illustrated in Fig.\ref{fig:ExperimentalSetup}(a), was our experimentally realized NN test bench.
Figure \ref{fig:ExperimentalSetup}(b) schematically depicts the experiment.
An optical plane wave $E^0$ illuminates the SLM's pixels, and the reflected field is filtered by a polarizing beam splitter (PBS).
The SLM combined with the PBS creates a $\cos(\cdot)$ nonlinearity and the SLM's pixels physically embody the neural network's state.
A quarter wave plate located between the PBS and the mirror directs the signal towards a camera, and a double pass through the diffractive optical element (DOE) establishes the recurrent connections $\mathbf{W}^{\textrm{DOE}}$ \cite{Brunner2015,Bueno2018,Maktoobi2020}.
Camera state $\hat{x}_{i}^{\textrm{cam}}(n)$ at integer time $n$ 
\begin{equation*}
    \hat{x}_{i}^{\textrm{cam}}(n) = \alpha \left| \sum_{j}^N W_{i,j}^{\textrm{DOE}} E_{j}(n)\right|^{2} \label{eq:xCam}
\end{equation*}  
\noindent is combined with external input information $u(n+1)$ and sent to the SLM, creating the network's state according to
\begin{equation}
\begin{array}{c}
\hat{x}_{i}(n+1)= \alpha |E_{i}^{0}|^{2} \\
\cos^{2} \Bigg[ \beta \cdot x_{i}^{\textrm{cam}}(n) + \gamma W_{i}^{\textrm{inj}} u(n+1)+\theta_{i} \Bigg]. \label{eq:x(n+1)}
\end{array}
\end{equation}
\noindent Here, $N$ is the recurrent layer's number of nodes, $\beta$ the feedback gain, $\gamma$ the input injection gain and $\alpha$ a normalization parameter.
The optical electric field and the nonlinearity's bias offset for node $i$ are $E_i^0$ and $\theta_i$, respectively.
Input information $u(n+1)$ is injected into the system according to random connections $\mathbf{W}^{\textrm{inj}}$.
As the network is constructed of physical neurons it harbours noise, which can either be additive or multiplicative, as well as correlated or uncorrelated.
More details about the theoretical treatment and propagation of noise in NNs can be found in \cite{Semenova2019}.

The polarization reflected by the PBS is imaged onto the DMD, whose mirrors are programmed to fixed angles of $\pm 12^{\circ}$ from normal incidence.
A photodiode only detects optical signals reflected off mirrors with $+12^{\circ}$ and implements Boolean readout weight matrix $W^{\textrm{DMD}}_{i=1\dots N}(k)$.
The RC's output is
\begin{equation}
\begin{array}{c}
y^{out}(k,n+1)\propto \\
\left|\sum_{i}^N W_{i}^{\textrm{DMD}}(k) (E_{i}^{0}-E_{i}(n+1))\right|^{2}\\
\propto \left| \sum_{i}^N W_{i}^{\textrm{DMD}}(k) \tilde{x}_{i}(n+1)\right|^{2}.\label{eq:Yout}
\end{array}
\end{equation}
\noindent Here, $k$ is the learning epoch and $\tilde{x}_{i}$ is the optical field of node $i$ arriving at the detector.
As in RC, we restrict learning to the optimization of the readout weights.
Finally, the absence of negative weights is partially mitigated by distributing the offset phases $\theta_i |_{i=1 \cdots N}$ randomly between $\theta_{0}+\delta\theta_i$ and $\theta_{0}+\Delta\theta+\delta\theta_i$, where $\delta\theta_i$ is a random Gaussian distribution \cite{Bueno2018}.
Internal $\mathbf{W}^{\textrm{DOE}}$ and readout $\mathbf{W}^{\textrm{DMD}}(k)$ connections are therefore realized in passive and fully parallel photonic hardware.

\section{Boolean evolutionary learning}
\label{sec:greedyLearning}

Must current learning techniques require complete knowledge of the network's state \cite{Jaeger2004}, all connection weights and potentially all gradients \cite{LeCun2015}.
In a hardware network this demands probing (and most probably externally storing) the value of each node and connection, which necessitates auxiliary circuitry of a complexity potentially exceeding the actual neural network.
This jeopardizes precisely the benefits one targets when mapping a neural network onto hardware.
We therefore employ learning that only tracks the computation error's evolution, and hence imposes no constraint on the type of neurons, and more broadly, on hidden layers as a whole.
Such an implementation's complexity therefore does not depend on, and hence does not limit the NN's size.

Here, we optimize the DMD's configuration simply by measuring the impact of mirror modifications onto computing error $\epsilon(k)$, one mirror at a time.
The objective is to modify $\mathbf{W}^{\textrm{DMD}}(k)$ during the $k= 1,2, ...,K$ learning epochs such that output $y^{\textrm{out}}(n + 1) $ best approximates target $\mathcal{T}(n+1)$.
Our Boolean learning algorithm can be divided into three conceptual sections:

\vspace{0.25cm}
\begin{description}
    \item[I. Mutation]
\end{description}

\begin{eqnarray}
\mathbf{W}^{\textrm{select}}(k) = \textrm{rand}(N)\cdot \mathbf{W}^{\textrm{bias}}(k) , \label{eq:Wselect} \\
l(k) = \textrm{max}(\mathbf{W}^{\textrm{select}}(k)) , \label{eq:lk} \\
W^{\textrm{DMD}}_{l(k)}(k+1) = \neg (W^{\textrm{DMD}}_{l(k)}(k)) , \label{eq:WDMDmodifExploration} \\
\mathbf{W}^{\textrm{bias}}(k+1) = 1/N + \mathbf{W}^{\textrm{bias}}(k),  W^{\textrm{bias}}_{l(k)}=0. \label{eq:Wbias}
\end{eqnarray}

We create a vector with $N$ independent and identically distributed random elements between 0 and 1 ($\textrm{rand}(N)$), and $\mathbf{W}^{\textrm{bias}}$ offers the possibility to modifying the otherwise stochastic selector $\mathbf{W}^{\textrm{select}}(k) \in \mathbb{R}^N$, Eq. \eqref{eq:Wselect}.
The largest entry's position in $\mathbf{W}^{\textrm{select}}(k)$ is $l(k)$, Eq. \eqref{eq:lk}, which determines the Boolean readout weight $W^{\textrm{DMD}}_{l(k)}(k)$ to be mutated via a logical inversion (operator $\neg(\cdot)$), see Eq. \eqref{eq:WDMDmodifExploration}.

A fully stochastic Markovian descent is obtained with $\mathbf{W}^{bias}=\mathbb{1}$ and excluding Eq. \eqref{eq:Wbias}.
However, we also investigate exploration which avoids mutating a particular connection in near succession.
There, $\mathbf{W}^{bias}$ is randomly initialized at $k=1$, and at each epoch Eq. \eqref{eq:Wbias} increases the bias of all connections by $1/N$, while the currently modified connection's bias is set to zero.
The probability of again probing a particular weight reaches unity only after $N$ learning epochs have passed, and we therefore refer to this biased descent as \emph{greedy} learning.

\vspace{0.25cm}
\begin{description}
    \item[2. Error and reward signals]
\end{description} 

\begin{eqnarray}
\epsilon(k) = \frac{1}{T} \sum_{n=1}^{T}\left(\mathcal{T}(n+1)-\tilde{y}^{\textrm{out}}(k,n+1)\right)^{2},\label{eq:epsilon}  \\
r(k) = \left\{ \begin{array}{r c l}
1 \quad \textrm{if}\ \Delta\epsilon(k) < 0\\
0 \quad \textrm{if} \ \Delta\epsilon(k) \ge 0
\end{array} \right. \label{eq:r(k)} \\
\epsilon^{\textrm{min}}(k) = (1-r(k))\epsilon(k-1)+r(k)\epsilon(k) , \label{eq:epsilonMin} \\
k^{\textrm{min}} = (1-r(k))k^{\textrm{min}}+r(k)k. \label{eq:kMin}
\end{eqnarray}

Mean square error $\epsilon(k)$ is obtained from a sequence of $T$ data points according to Eq. \eqref{eq:epsilon}, and comparison to the previous error assigns a reward $r(k)=1$ only if a modification $\Delta\epsilon(k)=\epsilon(k) - \epsilon(k-1)$ was beneficial, Eq. \eqref{eq:r(k)}.
In that case the minimum error $\epsilon^{min}_{k}$ and the best learning epoch $k^{\textrm{min}}$ are updated, Eqs. \eqref{eq:epsilonMin} and \eqref{eq:kMin}.

\vspace{0.25cm}
\begin{description}   
    \item[3. Descent action]
\end{description}

\begin{equation}\label{eq:WDMDmodifReward}
W^{DMD}_{l(k),k} = r(k)W^{DMD}_{l(k),k}+(1-r(k))W^{DMD}_{l(k),k-1} \ .
\end{equation}
\noindent Based on reward $r(k)$, the DMD's current configuration either accepts or rejects the previous modification, Eq. \eqref{eq:WDMDmodifReward}.
For a noise-less system reward $r(k)$ is therefore simply based on the gradient found at position $l(k)$, and we will refer to this hypothetical gradient of a noise-less system as the \emph{systematic} gradient. 
Simultaneously modifying groups of DMD mirrors is straight forward in principle, however, we found that convergence in our system and task is slower in that case.

\begin{figure}[t]
    \centering
    \includegraphics[width=0.95\columnwidth]{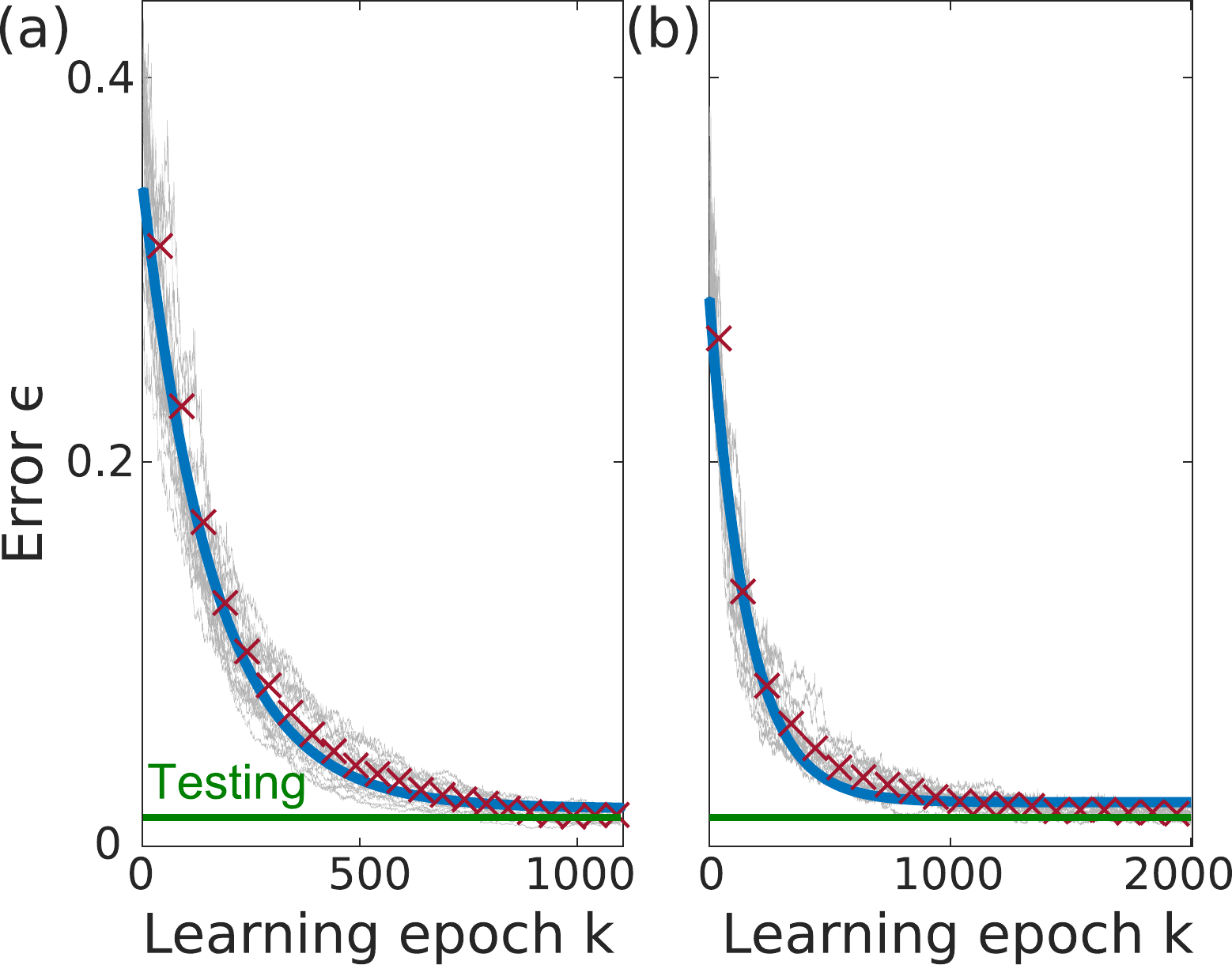}
    \caption{Learning performance, with individual (averaged) trajectories shown by gray data (red crosses).
        The green line is the testing error, and the blue is an exponential fit.
        Panel (a) and (b) were obtained with greedy and Markovian exploration, respectively.
    }\label{fig:20learningCurves}
    \hrule
\end{figure}

\section{Results} 
\label{sec:Results}

While such Boolean learning has been applied to a wide range of computational problems, recurrent neural networks have a particular relevance for dynamical signal processing, and we therefore explore one-step-ahead prediction of the chaotic Mackey-Glass sequence \cite{Jaeger2004,Bueno2016} with a Lyapunov exponent of $\sim 3 \cdot 10^{-3}$.
The chaotic sequence acting as input information $u(n+1)$ has zero mean and is normalized to its standard deviation, making error $\epsilon(k)$ the normalized mean square error.
Its first two hundred points are used as training signal $u(n + 1)$ and the target is $\mathcal{T}(n+1) = u(n + 2)$.
The first thirty time steps are removed due to their transient nature, and from the system's output $y(n+1)$ we subtract the mean and normalize by its standard deviation, creating output $\tilde{y}_{k}^{out}$ which is used in Eq. \eqref{eq:epsilon}.
Finally, the testing error is determined with an independent set of 9000 data-points unused in the training sequence.
Based on mutation of the readout weights, our concept explores an error landscape with height $\epsilon(k)$ and position $\mathbf{W}^{\textrm{DMD}}(k)$, and reward $r(k)$ drives the configuration from $\mathbf{W}^{\textrm{DMD}}(1)$ to a local minima at $\mathbf{W}^{\textrm{DMD}}(k^{\textrm{min}})$.
There our system will remain trapped due to an exploration step size of 1.
We will refer to one complete learning process for $k:1\rightarrow k^{\textrm{min}}$ as a \emph{minimizer}.

Understanding why generalization is possible for a training set size ($T=200-30$) not orders of magnitude larger than the number of to be optimized weights ($N=961$) is an interesting question.
Recent results on deep neural networks, triggered by the insightful analysis from \cite{belkin2018reconciling}, show that overparametrization may not preclude generalization.
See \cite{hastie2019surprises} for an account to this phenomenon using random matrix theory, starting from simple linear models and generalizing to kernel estimation.
In our setting we, however, might additionally postulate that we work below the overparametrization barrier due to the Boolean entries of $\mathbf{W}^{\textrm{DMD}}(k)$, which brings substantial rigidity into play.
The price one pays is making the problem harder from a computational optimization viewpoint \cite{Hadaeghi2019}.

Typically, the main metric for evaluating learning are speed of convergence $k^{\textrm{min}}$ and final inference error $\epsilon^{\textrm{min}}$.
However, in analogue neural hardware, reproducibility as well as robustness to noise and parameter drifts also play an essential role.
We start by collecting statistical information and measure 20 (14) curves for the greedy (Markovian) exploration.
All measurements started at an identical position $\mathbf{W}^{\textrm{DMD}}(1)$ and we therefore focus on the algorithm's exploration of the error-landscape.
Results are shown in Fig. \ref{fig:20learningCurves}, with individual learning curves as grey lines and their average as red crosses.
Panel (a) shows data for the greedy, panel (b) for Markovian exploration.
Convergence for both cases scales linear with network size $N$ \cite{Porte2019}, yet greedyness approximately divides $k^{\textrm{min}}$ compared to Markovian decent.

\subsection{Average and local features of convergence and minima}	
\label{subsec:Convergence}

On average, the error landscape topology excellently follows an exponential decay for both exploration strategies, see fit (blue line) to the average error (red crosses).
Comparing individual trajectories, however, reveals strong inter-trial differences significantly exceeding the noise level.
This diversity corresponds to the error landscape's topological richness probed by the different random descents, and trajectories range from rather smooth descents to paths including steep drops.
No correlation between the starting $\epsilon(1)$ and best performance $\epsilon^{min}$ was found, and for the many different minimizers our system never got stuck in a local minima with bad performance.
Crucially, the system's testing error $\epsilon^{\textrm{test}} = 15\cdot 10^{-3}$ excellently matches its training error $\epsilon^{\textrm{train}} = (14.2\pm1.5)\cdot 10^{-3}$, hence ruling out over fitting.
This data was obtained with greedy exploration, yet the agreement is equally good in the Markovian exploration case.

\begin{figure}[!ht]
    \centering
    \includegraphics[width=1\columnwidth]{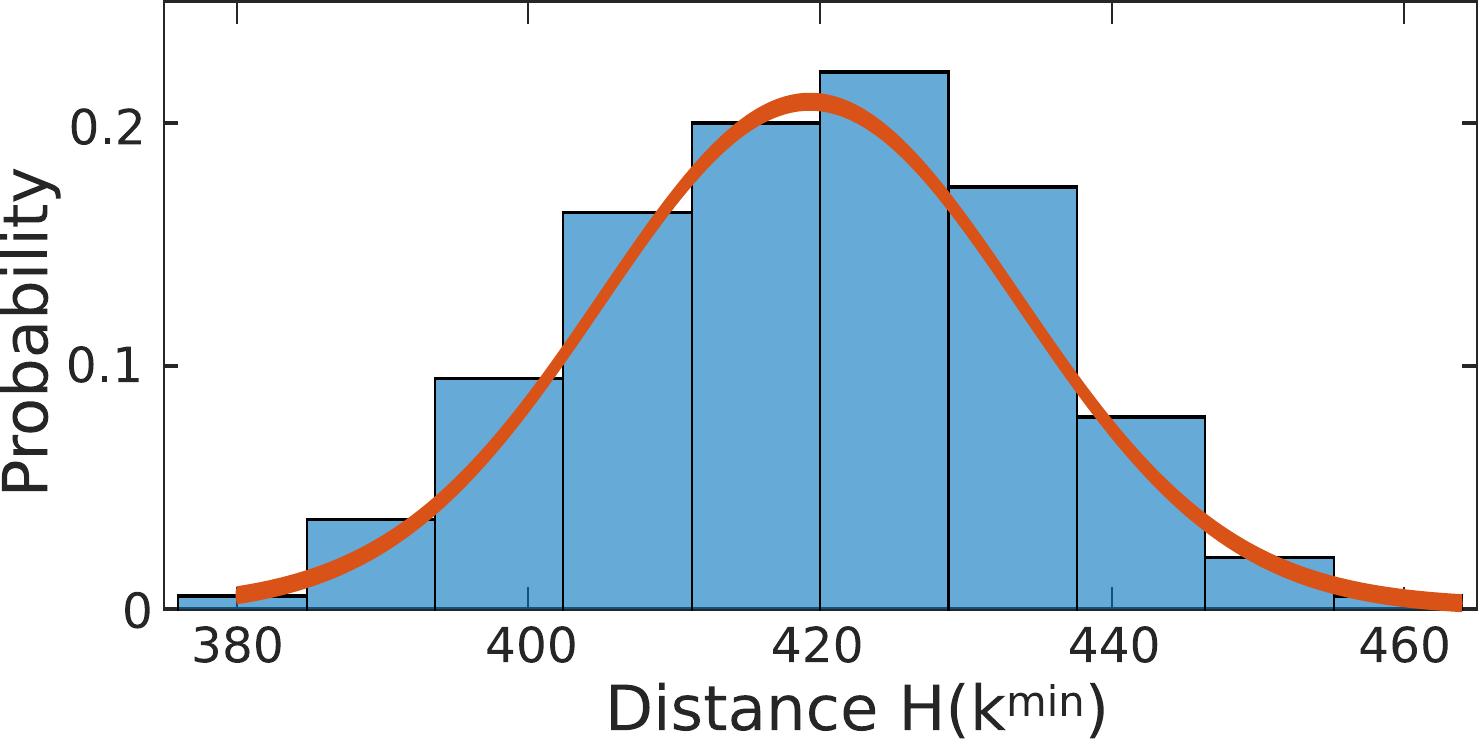}
    \caption{Probability distribution of the Hamming distances at $k=N$, obtained from the different greedy learning curves displayed in Fig. \ref{fig:20learningCurves}(a).
        Red curve is a Gaussian fit which is centered at $419$ and has a standard deviation of $29$.}\label{fig:190HammingDistanceAnd3LC}
    \hrule
\end{figure}

Nevertheless, despite the similar $\epsilon^{\textrm{min}}$, we find that optimal DMD configurations of individual learning trials have negligible correlation between each other.
All minimizers therefore arrive at different local minima, and we encounter a surprising regularity in their geometric arrangement.
The separation between two Boolean readout configurations $\mathbf{W}^{\textrm{DMD,a}}(k)$ and $\mathbf{W}^{\textrm{DMD,b}}(k)$ is determined by Hamming distance $H(k^{\textrm{a}},k^{\textrm{b}})=\sum_i | W^{\textrm{DMD,a}}_{i}(k^{\textrm{a}}) -W^{\textrm{DMD,b}}_{i}(k^{\textrm{b}}) |$.
For the 20 minimizers we obtain $20\cdot (20-1) / 2 = 190$ distances between their respective minima, and their statistical distribution obtained for greedy exploration is shown in Fig. \ref{fig:190HammingDistanceAnd3LC}.
The red line is a Gaussian fit centred at $\bar{H}=419$ and with a half width at $1/e$ of 14.
Data shows a very specific and unusual error landscape topology:
local minima appear not to be irregularly distributed, nor located in a particular region.
Instead, the negligible inter-minima correlations and systematic, non-dispersed inter-minima distances reveal a uniform distribution in an quasi periodic arrangement.
Again, we find that Markovian exploration results in an identical behavior.

\section{Noise sensitivity}
\label{subsec:noise}

\begin{figure}[h!]
    \centering
    \includegraphics[width=0.95\columnwidth]{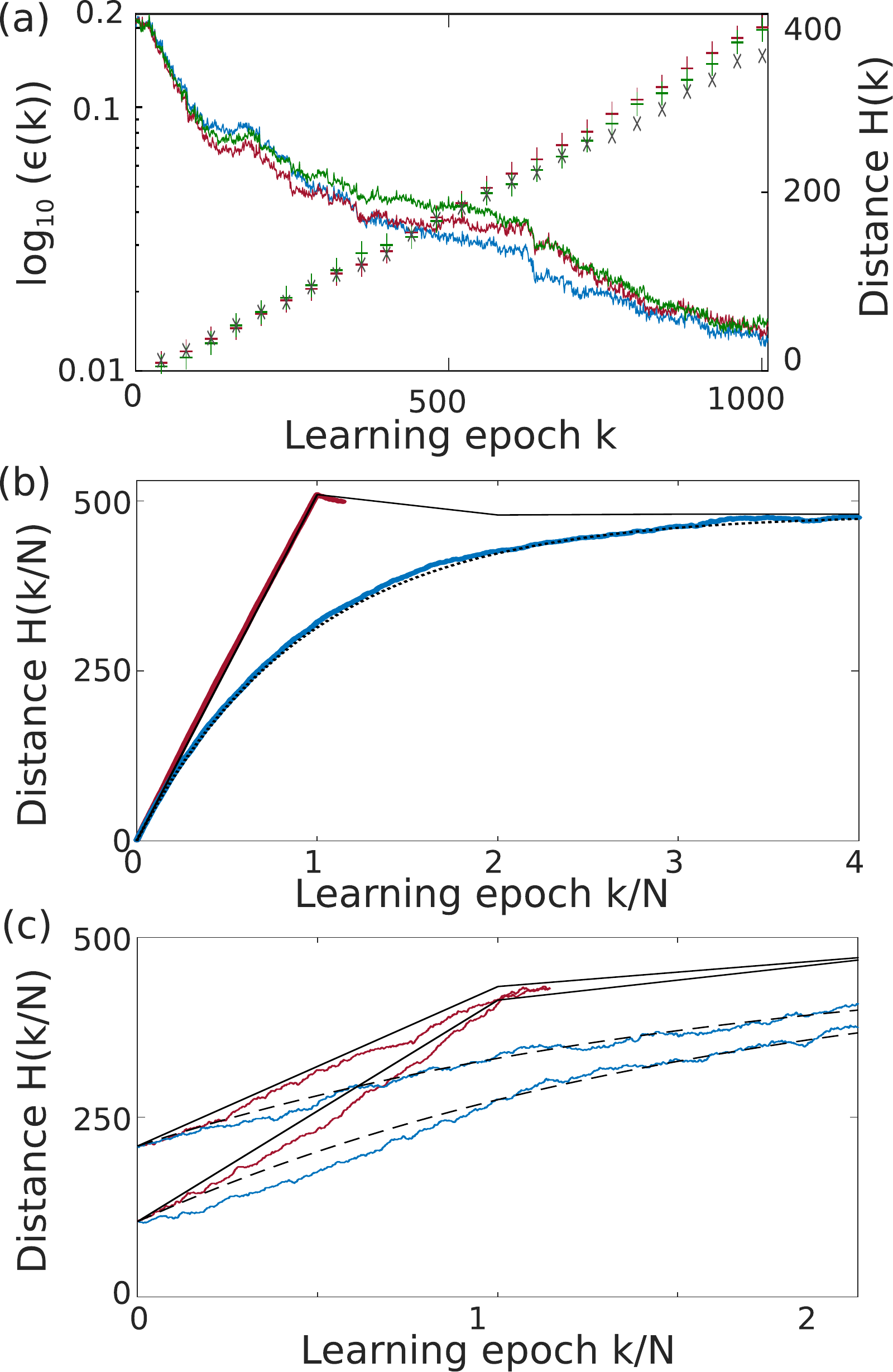}
    \caption{(a) Three minimizers starting from identical position are measured in parallel.
        The red and green (slaves) optimization paths test the dimension $l(k)$ determined by the blue minimizer (master), reward of each mutation are evaluated and applied individually.
        The lines show the individual errors on a logarithmic scale, crosses depict the Hamming distance's evolution between the three minimizers.
        (b) Hamming distance evolution with learning epoch $k$ normalized by the network's Size $N$.
        Random mutation (blue line) leads to a smooth saturation function behaviour, while biased (greedy) mutation (red dashed line) results in linear intervals of length $N$.
        (c) The same characteristics are found for two minimizers starting already separated by a distance $(1)\neq0$, and the behaviour is therefore generally true for Boolean learning in noisy hardware.}\label{fig:HammingEvol}
    \hrule
\end{figure}

To further investigate this phenomena, we reduce the number of  uncertainties in the system's descent through the error landscape.
We measure three optimization paths starting at the same $\mathbf{W}^{DMD}_{1}$.
One of the three minimizers acts as a master and defines mutation sequence $l(k),k\in [1\dots K]$, which the other two minimizers follow as slaves.
Crucially, all three compute their own rewards $r(k)$ and hence independently evaluate the same mutations.
Keeping the potentially systematic error of a slow experimental parameter drift in mind, the three systems are evaluated at each learning epoch $k$ before advancing to $k+1$.
A single minimizer takes $\sim$ 20 hours, and sequential evaluation would amplify susceptibility to slow experimental parameter drifts which are taking place on the scale of hours in our experiment.

Results are shown in Fig. \ref{fig:HammingEvol}(a).
The blue, green and red lines correspond to the different errors $\epsilon(k)$, plotted on a semi-logarithmic scale for the master and two slaves, respectively.
The different data have a high degree of similarity with an average correlation of 99.4\%, yet locally one can identify some significant differences.
We computed the temporal evolution of the Hamming distances $H(k)$ of the two slaves to their master (red and green crosses) and between the two slaves (gray crosses); all three grow linearly at essentially the same rate.
Without noise, each reward $r(k)$ would be identical for each minimizer, who would consequently all follow the same trajectories $\mathbf{W}^{\textrm{DMD}}(k)$ arriving at the same minima $\mathbf{W}^{\textrm{DMD}}(k^{\textrm{min}})$.	

We therefore have to consider the impact of noise upon the system's error $\epsilon(k)$.
The response of error $\epsilon(k)$ to a modification in the system's output $\Delta y^{\textrm{out}}(k)$ is 
\begin{equation}\label{eq:errorGrad}
\Delta\epsilon(k) = \dot{\epsilon}(k)\cdot \Delta y^{\textrm{out}}(k)
\end{equation}
\noindent in general, and $\Delta y^{\textrm{out}}(k)$ can be induced either by a modification to $\mathbf{W}^{\textrm{DMD}}$, or by noise.
Error $\epsilon(k)$ therefore has always the same sensitivity $\dot{\epsilon}(k)$ towards noise and weight inversions.
This symmetry is important, and some general considerations serve as guidance in the interpretation of Eq. \ref{eq:errorGrad}.
The amplitude of all network nodes $\tilde{x}_i$ are Gaussian distributed due to the SLM's illumination by a collimated Gaussian beam.
Randomly changing one readout weight results in $\Delta y^{\textrm{out}}(k)$ according to a normalized Gaussian distribution with a width of $\Delta y^{\textrm{out,learn}}(k)=\sigma^{l}(k)$.
We have carefully characterized the noise of all elements in our opto-electronic NN, and its impact upon $y^{\textrm{out}}(k)$ is excellently approximated by a normalized Gaussian white noise distribution with a width of $\Delta y^{\textrm{out,noise}}(k)=\sigma^{n}(k)$ \cite{Semenova2019}.
Readout weights $\mathbf{W}^{\textrm{DMD}}$ remain approximately evenly distributed between zeros and ones for all $k$, and we recall that learning does not modify $\tilde{x}_i$.
We can therefore assume that modifications to $y^{\textrm{out}}$ induced by learning and noise remain constant for all $k$, hence $\sigma^{l}(k)=\sigma^{l}$ and $\sigma^{n}(k)=\sigma^{n}$.

Noise and weight modifications have to be recognized as independent players, whose action upon learning is somehow competitive.
The objective of modifying a readout weight is to probe the error landscape's systematic gradient, which, however, is contaminated by noise which can potentially exceed the systematic gradient in the opposite direction.
The consequence is a change in the sign of $\Delta\epsilon(k)$, in which case reward $r(k)$ is inverted.
Equation \eqref{eq:errorGrad} shows that the impact of noise and weight modifications scale equally at each $k$, and constant $C$ is the probability that noise modified reward $r(k)$, $1-C$ that the reward is kept in accordance with the systematic gradient.
The analytical derivation of $C$ is possible, yet beyond the scope of this manuscript.

Probability $C$ is the driving force behind the growing separation between two identical minimizers, and two situations are relevant.
The first situation occurs when $r(k)$ for one minimizer is inverted by noise while the other preserves its systematic value, which has a probability of $C(1-C) + (1-C)C = 2C(1-C)$.
The other situation is if both minimizers have an identical reward $r(k)$, which can either be the consequence of both retaining their systematic result, or for both being inverted by noise, with a combined probability of $(1-C)^2+C^2=1-2C(1-C)$.
The first situation leads to $H(k+1)\neq H(k)$, the second to $H(k+1)=H(k)$, and the Hamming distance's rate equation is
\begin{equation}
\begin{array}{c}
\Delta H(k+1) = \rho^{\textrm{id}}(k+1)2C(1-C)\\
 - \rho^{\textrm{op}}(k+1)2C(1-C).\label{eq:HamingRateEqu}
\end{array}
\end{equation}
\noindent Here, $\rho^{\textrm{id}}(k)$ and $\rho^{\textrm{op}}(k)$ are the probability of finding both minimizers' weights $l(k)$ to be identical or opposite, respectively.
Using $\rho^{\textrm{id}}(k)= 1 - \rho^{\textrm{op}}(k)$, we arrive at
\begin{eqnarray}
H(k+1) = H(k) + \tilde{C}(1-2 \rho^{\textrm{op}}(k)),\\  \label{eq:HammingEvol}
\Delta H(k+1) = \tilde{C}(1-2 \rho^{\textrm{op}}(k)), \label{eq:HammingGradient}
\end{eqnarray}
\noindent where $\tilde{C} = 2C(1-C)$.

The Hamming distance's evolution is therefore governed by noise and by how the selection procedure picks weight $W^{\textrm{DMD}}_{l(k)}(k)$ from a population with a certain $\rho^{\textrm{\textrm{op}}}(k)$.
For fully random mutation, the probability of a weight to be selected is identical at every $k$, and hence the Hamming distance at the previous epoch $k$ determines the probability of two weights being opposite in their configuration: $\rho^{\textrm{op}}(k) = H(k)/N$.
For greedy learning, the bias term in Eq. \eqref{eq:Wselect} causes mutations hardly ever to repeat the same weight during an interval specified by $k= k'+ aN, k'\in[1, N]$, with non-negative integer $a$.
The probability of both minimizers to be configured opposite for all $k'$ and a specific $a$ is therefore their Hamming distance at the end of the previous interval: $\rho^{\textrm{op}}(k) = H(aN)/N$.
Interestingly, this results in constant slopes $\Delta H(k'+aN)$ for each $k'$.

Figure \ref{fig:HammingEvol}(b) shows the evolution of Hamming distance $H(k/N)$, and since all minimizers start at the same position $\mathbf{W}^{\textrm{DMD}}(1)$ we always have $H(1)=0$.
Greedy mutations in the experiment (analytics) are the red line (black dashed line), while random mutations in the experiment (analytics) are the blue line (black solid line).
For both scenarios, greedy and random descent, the experimental data is the average obtained from 20 minimizers.
We then changed the starting conditions and realized two parallel minimizers which started with a separation $H(1)>0$, see Fig. \ref{fig:HammingEvol}(c).
In general, the evolution according to Eq. \eqref{eq:HammingGradient} perfectly reproduces results of the highly different experimental learning scenarios, in particularly for the averaged data.

\begin{figure}[t]
    \centering
    \includegraphics[width=0.95\columnwidth]{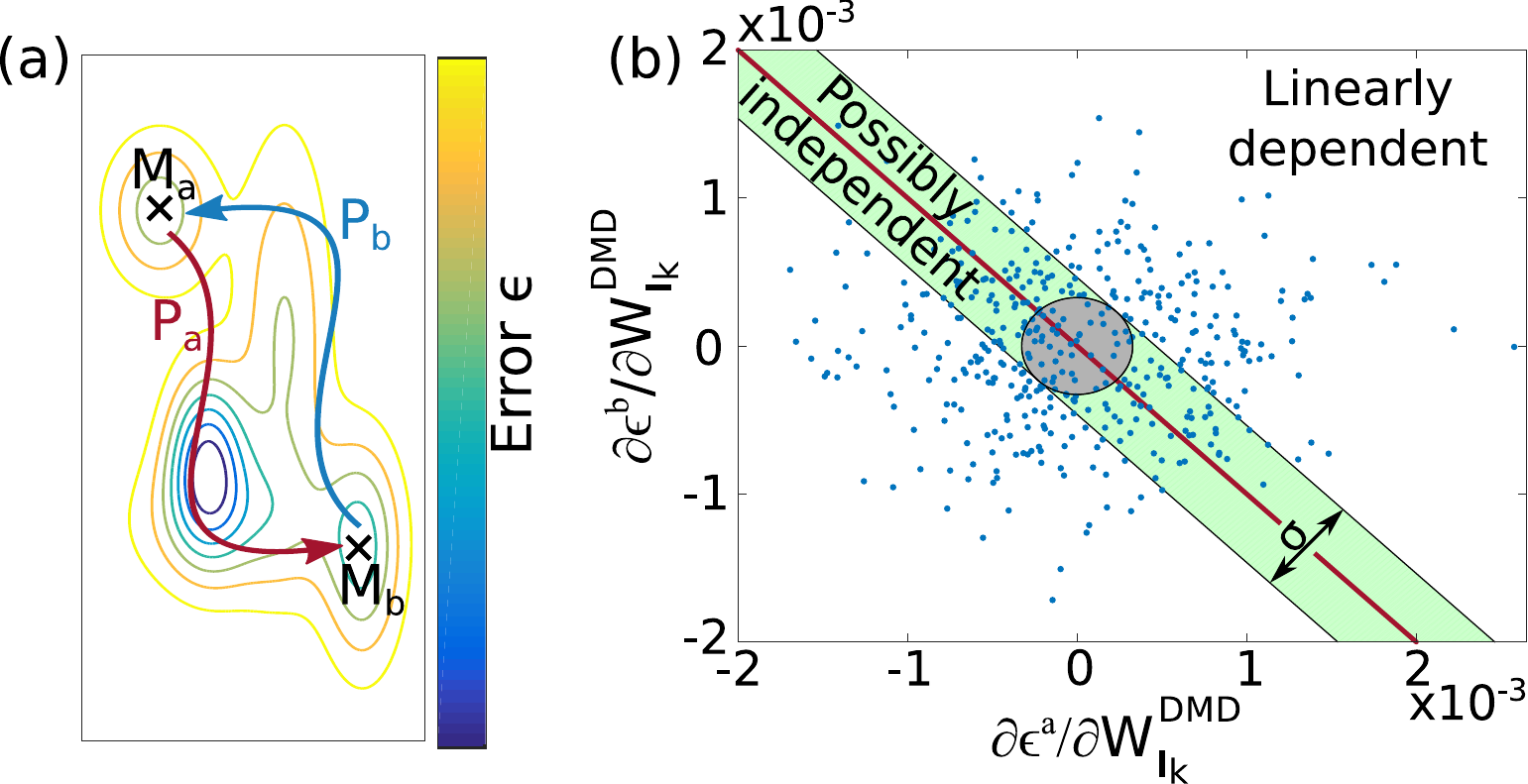}
    \caption{(a) Two inverted paths are probed between two local minima.
        (b) Error gradient for all readout weights encountered along path one and two as x and y axis, respectively.
        The red diagonal corresponds to linear independent weights, and the uncertainty induced by noise is indicated by the green area.
        Within, weights are potentially linear independent.
        Outside the green area weights are linearly dependent, and for data inside the grey circle of diameter $\sigma$ no classification is possible.}\label{fig:LinDepend}
    \hrule
\end{figure}

Different minimizers therefore always arrive at final readout configurations which share no common feature.
This suggests a closer look into the role and relevance of individual weights: how many induce a systematic contribution to convergence at all, and if their gradients depend on the sequence of previous mutations.
We optimized readout weights via two minimizers starting at different random positions $\mathbf{W}^{\textrm{DMD,a}}(1)$ and $\mathbf{W}^{\textrm{DMD,b}}(1)$, which arrived at two distinct local minima $M_{a}=\mathbf{W}^{\textrm{DMD,a}}(k^{\textrm{min,a}})$ and $M_{b}=\mathbf{W}^{\textrm{DMD,b}}(k^{\textrm{min,b}})$.
Once there, we determined the list of $m$ weights where $M_b$ differs from $M_a$.
The list is randomly arranged in sequence $\mathbf{l} \in[l(1),l(2), \dots, l(m-1), l(m)]$ according to which we invert the corresponding weights $W_{l(k)}^{\textrm{DMD,a}}(k)$ and $W_{l(k)}^{\textrm{DMD,b}}(k)$ for $k\in[1,m]$.
Importantly, this mutation is always kept and no optimization based on reward $r(k)$ is taking place.
Starting from $M_a$ ($M_b$), this results in a random path $P_{a}: M_a \longrightarrow M_b$ ($P_{b}: M_b \longrightarrow M_a$).
As the weights addressed in sequence $\mathbf{l}(k)$ are the ones in an opposite configuration for $M_a$ and $M_b$, $P_{a}$ and $P_{b}$ connect both minima along inverted trajectories, see Fig. \ref{fig:LinDepend}(a).
We probe error $\epsilon(k)$ along $P_{a}$ and $P_{b}$ and determine error gradients $\dot{\epsilon}^{a}(k)=\epsilon^{a}(k)-\epsilon^{a}(k^{\textrm{min,a}}$ and $\dot{\epsilon}^{b}(k)=\epsilon^{b}(k)-\epsilon^{b}(k^{\textrm{min,b}}$.
Results are shown in Fig. \ref{fig:LinDepend}(b) in the $(\dot{\epsilon}_a, \dot{\epsilon}_b)$-plane, and for this experiment we obtained $m=430$ different dimensions between $M_a$ and $M_b$.
Only $\approx$11\% of the 430 gradients consistently remained below our system's noise floor $\dot{\epsilon}\sigma^{n}$, indicated by the gray circle.

Weights insensitive (sensitive) to preceding optimizations correspond to linearly independent (linearly dependent) NN dimensions.
Linearly independent NN dimensions must always induce the same gradient, regardless of the previous optimization path, and linear independent dimensions therefore have to be located on the red diagonal line in Fig. \ref{fig:LinDepend}(b).
The Figure's green area indicates the linearly independent criteria when considering the impact of noise $\sigma^{n}$, and we find $\approx$ 30\% of 430 dimensions fall into this category.
However, this is only a necessary criterion; a sufficient criteria requires allocating dimensions inside this area for all potential configurations of the remaining $m-1$ dimensions, which for the $2^{429}$ possibilities is impossible.
The NN dimensions whose weight configuration depends on the previously optimized weights lie outside the gray and green areas.
This is a sufficient criteria for linear dependent NN dimension, and we find $\approx$ 59\% of 430 to be contained inside this set.
There appears to be no concentration of potentially linearly independent dimensions towards the red diagonal line, and hence we conclude that these also mostly belong to the set of linear dependent dimensions.

\section{\label{Discussion}Discussion}

Our experimental findings and analytical descriptions are the first of their kind and stimulate a fundamental discussion.
Equation \eqref{eq:errorGrad} is of interesting consequence for noisy hardware NNs comprising linear readout weights.
It links the susceptibility of $\Delta\epsilon(k)$ to NN noise to the system's location inside the error landscape $\epsilon(k)$.
Experimentally we obtained noise induced variations $\dot{\epsilon}(150)\sigma^{n}=4.4\cdot 10^{-3}$ and $\dot{\epsilon}(961)\sigma^{n}=0.6 \cdot 10^{-3}$.
Our learning curves excellently agree with exponential convergence $\epsilon(k)=\epsilon^0 e^{-\alpha k}$, for which $\dot{\epsilon}(k)\propto \epsilon(k)$.
Error and gradient therefore evolve in a linearly proportional manner, and  $\epsilon(150)/\epsilon(961)=9.8$ in close agreement with the noise sensitivity's evolution $\dot{\epsilon}(150)\sigma^{n}/\dot{\epsilon}(961)\sigma^{n}\approx 7.3$ confirms this fundamental relationship.
Alternative noise-mitigation approaches are a derivative of these findings.
Simply suppressing noise on a hardware level is potentially expensive, and topological requirements can limit mitigation based on connectivity statistics \cite{Semenova2019}.
One might therefore curb the impact of noise by modified learning strategies, for example by amending an optimization's cost function by the gradients around a minima, hence reducing $\dot{\epsilon}$ in the minima's neighbourhood.

Equation \eqref{eq:HammingGradient} shows that for $C>0$ the Hamming distance between readout weights of two systems will always tend towards complete decorrelation as $H(k)|_{k\rightarrow\inf}=N/2$.
Even for 100 \% identical networks one will therefore never obtain similar readout configurations \cite{Freiberger2019}.
This finding can most likely be extended to the non-Boolean case and to the weights between layers of analogue deep NNs.
The field of learning implemented in physical and hence noisy substrates is only in its infancy \cite{Bueno2018,Antonik2017,Shen2016}, and confirmation of our findings in other hardware systems would prove the generality of our result.
Finally, the human brain is a very noisy network indeed \cite{Suarez-Perez2018}, and our findings have interesting implications for the field of theoretical neuroscience.

We have shown that the large majority of our NN's dimensions are most likely linear dependent. What this means in pratical terms is that each modification of a weight has to be interpreted in the context of all previous modifications. Each configuration $\mathbf{W}^{\textrm{DMD}}(k)$ therefore encodes the history of modifications to the reward due to noise during the previous learning epoch.

One direct consequence for applications is that one cannot simply transfer or swap weight configurations between optimized analogue neural networks, even for potentially available identical twin networks.
The reason is that optimized configurations are not only the consequence of error landscape, system properties and noise, but also of the precise history of noise during an exploration path.
Even perfectly reproducible hardware networks will therefore always have to be individually trained for optimal performance; simply uploading a configuration will potentially not work.
A 'school' where each neural network learns individually might therefore be required.
Finally, our findings open a new field where such twin-minimizers could be considered for probing and interrogating unknown hardware neural networks.
The average divergences shown in Fig. \ref{fig:HammingEvol}(b) agree exceptionally well with our model, and based on such data one can therefore make accurate inferences about the noise properties of a hardware NN and about its error landscape exploration strategy.

\section{\label{subsec:conclusion}Conclusion}	

In our work we have investigated the intricate interactions between different learning concepts and the noise inherently present in analogue neural networks.
We experimentally showed that trajectories of individual minimizers strongly diverge, and were able to analytically link this divergence to a constant ration between output error and noise susceptibility.
Our analytical description only assumes a linear multiplication between a NN's state and its readout weights, and hence should be generally applicable to this wide class of analogue hardware NNs.


\vspace{0.5cm}

\section*{Funding Information}
The authors acknowledge the support of the Region Bourgogne Franche-Comt\'{e}. 
This work has received funding from the European Union’s Horizon 2020 research and innovation programme under the Marie Skłodowska-Curie grant agreements No 860830 (POST DIGITAL) and No 713694 (MULTIPLY).
This work was also supported by the EUR EIPHI program (Contract No. ANR-17-EURE- 0002), the BiPhoProc project (Contract No. ANR-14-OHRI- 0002-02), and by the Volkwagen Foundation (NeuroQNet). 

\bibliographystyle{ieeetr}
\bibliography{references}

\end{document}